\documentclass[10pt,twocolumn,letterpaper]{article}

\usepackage[camera-ready]{cvpr}      

\usepackage[dvipsnames]{xcolor}

\definecolor{cvprblue}{rgb}{0.21,0.49,0.74}
\usepackage[pagebackref,breaklinks,colorlinks,citecolor=cvprblue]{hyperref}

\usepackage{indentfirst}

\def\paperID{****} 
\def\confName{****}
\def\confYear{****}

\title{Animate Anyone: Consistent and Controllable Image-to-Video Synthesis for Character Animation}

\author{Li Hu,  Xin Gao,  Peng Zhang,  Ke Sun,  Bang Zhang,  Liefeng Bo\\
Institute for Intelligent Computing, Alibaba Group\\
\tt\small \{hooks.hl, zimu.gx, futian.zp, xisheng.sk, zhangbang.zb, liefeng.bo\}@alibaba-inc.com\\
\small \url{https://humanaigc.github.io/animate-anyone/}
}

\begin{document}

\twocolumn[{%
\renewcommand\twocolumn[1][]{#1}%
\maketitle
\begin{center}
    \centering
    \captionsetup{type=figure}
    \includegraphics[width=1.0\textwidth]{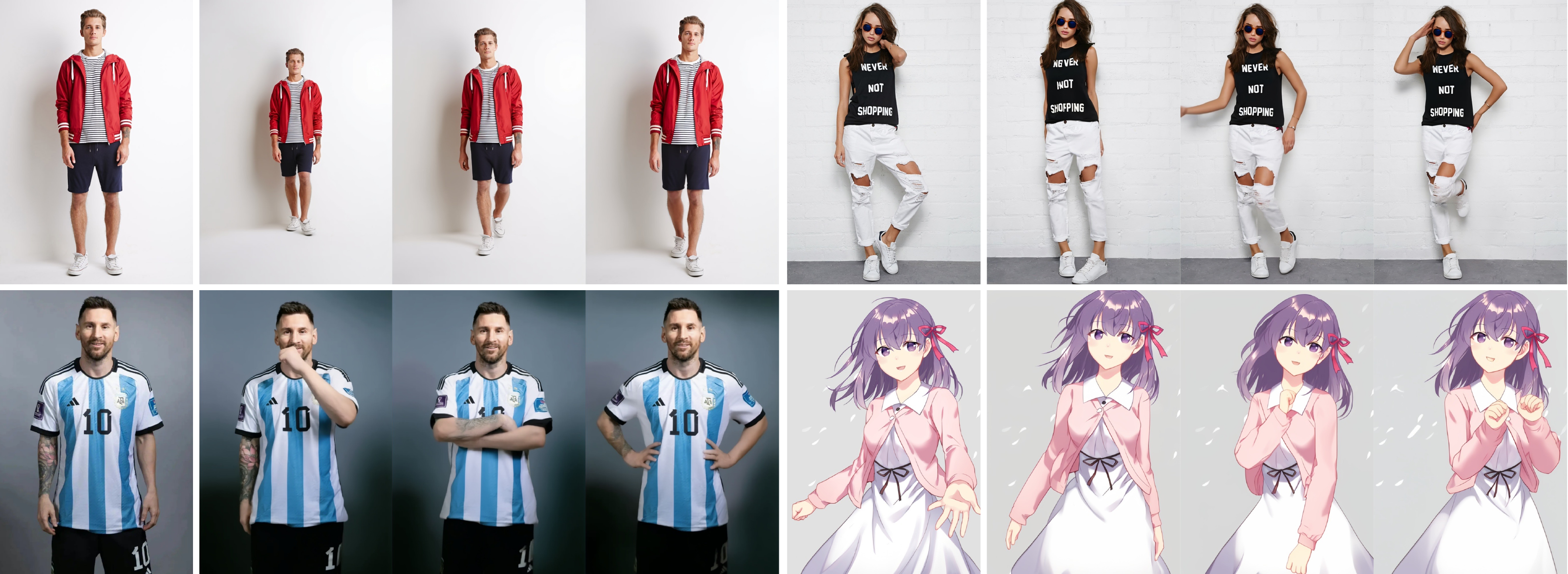}
    \captionof{figure}{Consistent and controllable character animation results given reference image (the leftmost image in each group) . Our approach is capable of animating arbitrary characters, generating clear and temporally stable video results while maintaining consistency with the appearance details of the reference character. }
    \label{fig:f1}
\end{center}%
}]


\maketitle

\begin{abstract}


Character Animation aims to generating character videos from still images through driving signals. Currently, diffusion models have become the mainstream in visual generation research, owing to their robust generative capabilities. However, challenges persist in the realm of image-to-video, especially in character animation, where temporally maintaining consistency with detailed information from character remains a formidable problem. In this paper, we leverage the power of diffusion models and propose a novel framework tailored for character animation. To preserve consistency of intricate appearance features from reference image, we design ReferenceNet to merge detail features via spatial attention. To ensure controllability and continuity, we introduce an efficient pose guider to direct character's movements and employ an effective temporal modeling approach to ensure smooth inter-frame transitions between video frames. By expanding the training data, our approach can animate arbitrary characters, yielding superior results in character animation compared to other image-to-video methods. Furthermore, we evaluate our method on image animation benchmarks, achieving state-of-the-art results.

\end{abstract}

\section{Introduction}

Character Animation is a task to animate source character images into realistic videos according to desired posture sequences, which has many potential applications such as online retail, entertainment videos, artistic creation and virtual character. Beginning with the advent of GANs\cite{gan,wgan,stylegan}, numerous studies have delved into the realms of image animation and pose transfer\cite{fomm,mraa,ren2020deep,tpsmm,siarohin2019animating,zhang2022exploring,bidirectionally,everybody}. However, the generated images or videos still exhibit issues such as local distortion, blurred details, semantic inconsistency, and temporal instability, which impede the widespread application of these methods.

In recent years, diffusion models\cite{denoising} have showcased their superiority in producing high-quality images and videos. Researchers have begun exploring human image-to-video tasks by leveraging the architecture of diffusion models and their pretrained robust generative capabilities. DreamPose\cite{dreampose} focuses on fashion image-to-video synthesis, extending Stable Diffusion\cite{ldm} and proposing an adaptar module to integrate CLIP\cite{clip} and VAE\cite{vae} features from images. However, DreamPose requires finetuning on input samples to ensure consistent results, leading to suboptimal operational efficiency. DisCo\cite{disco} explores human dance generation, similarly modifying Stable Diffusion, integrating character features through CLIP, and incorporating background features through ControlNet\cite{controlnet}. However, it exhibits deficiencies in preserving character details and suffers from inter-frame jitter issues. 

Furthermore, current research on character animation predominantly focuses on specific tasks and benchmarks, resulting in a limited generalization capability. 
Recently, benefiting from advancements in text-to-image research\cite{dalle2,glide,imagen,ldm,composer,ediffi}, video generation (e.g., text-to-video, video editing)\cite{animatediff,cogvideo,fatezero,imagenvideo,text2videozero,tuneavideo,videocomposer,align,gen1,makeavideo,vdm} has also achieved notable progress in terms of visual quality and diversity.
Several studies extend text-to-video methodologies to image-to-video\cite{videocomposer,videocrafter1,i2vgen,animatediff}. 
However, these methods fall short of capturing intricate details from images, providing more diversity but lacking precision, particularly when applied to character animation, leading to temporal variations in the fine-grained details of the character's appearance. Moreover, when dealing with substantial character movements, these approaches struggle to generate a consistently stable and continuous process. 
Currently, there is no observed character animation method that simultaneously achieves generalizability and consistency.

In this paper, we present \textit{Animate Anyone}, a method capable of transforming character images into animated videos controlled by desired pose sequences. 
We inherit the network design and pretrained weights from Stable Diffusion (SD) and modify the denoising UNet\cite{unet} to accommodate multi-frame inputs.
To address the challenge of maintaining appearance consistency, we introduce ReferenceNet, specifically designed as a symmetrical UNet structure to capture spatial details of the reference image. At each corresponding layer of the UNet blocks, we integrate features from ReferenceNet into the denoising UNet using spatial-attention\cite{attention}. This architecture enables the model to comprehensively learn the relationship with the reference image in a consistent feature space, which significantly contributes to the improvement of appearance details preservation. 
To ensure pose controllability, we devise a lightweight pose guider to efficiently integrate pose control signals into the denoising process. 
For temporal stability, we introduce temporal layer to model relationships across multiple frames, which preserves high-resolution details in visual quality while simulating a continuous and smooth temporal motion process.

Our model is trained on an internal dataset of 5K character video clips. Fig.~\ref{fig:f1} shows the animation results for various characters. Compared to previous methods, our approach presents several notable advantages. 
Firstly, it effectively maintains the spatial and temporal consistency of character appearance in videos. Secondly, it produces high-definition videos without issues such as temporal jitter or flickering. Thirdly, it is capable of animating any character image into a video, unconstrained by specific domains. 
We evaluate our method on three specific human video synthesis benchmarks (UBC fashion video dataset\cite{dwnet}, TikTok dataset\cite{tiktok} and Ted-talk dataset\cite{mraa}), using only the corresponding training datasets for each benchmark in the experiments. Our approach achieves state-of-the-art results.
We also compare our method with general image-to-video approaches trained on large-scale data and our approach demonstrates superior capabilities in character animation. 
We envision that \textit{Animate Anyone} could serve as a foundational solution for character video creation, inspiring the development of more innovative and creative applications.

\begin{figure*}[!t]
\begin{center}
	\setlength{\fboxrule}{0pt}
	\fbox{\includegraphics[width=0.99\textwidth]{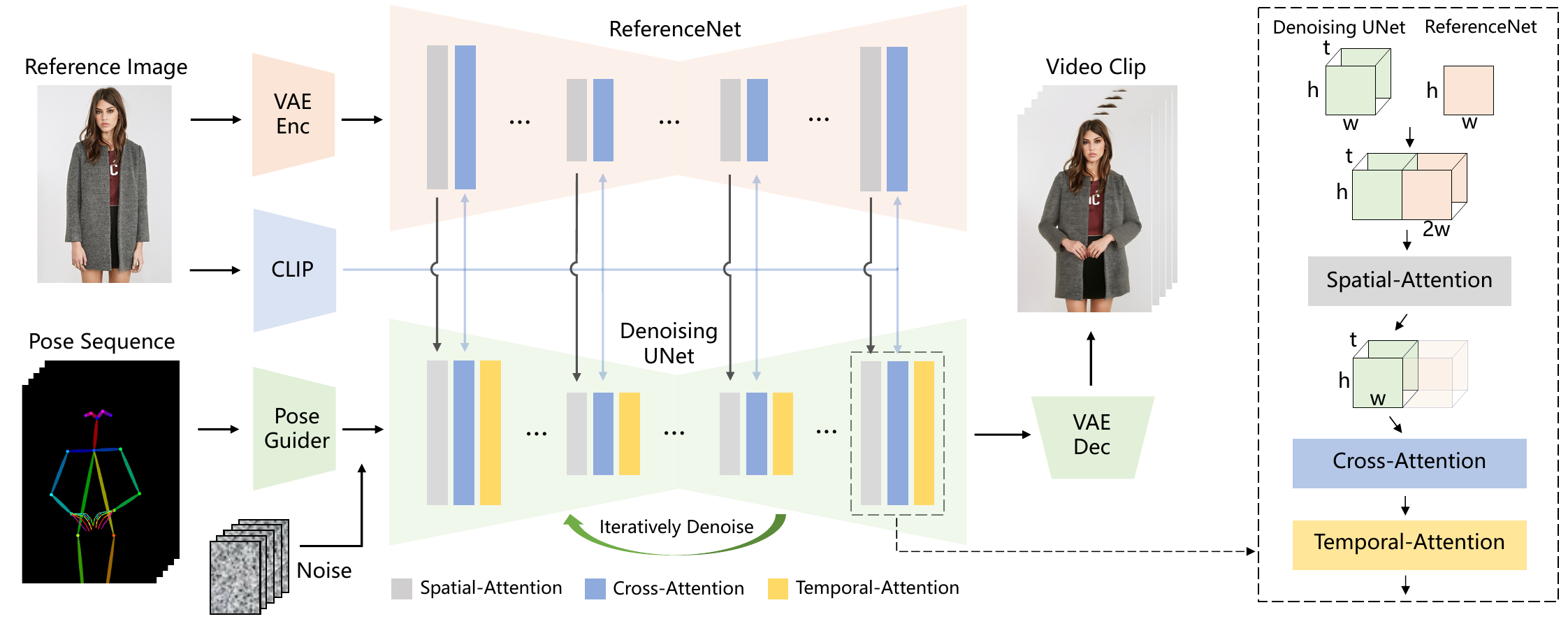}}
\end{center}
\vspace{-0.5cm}
\caption{The overview of our method. The pose sequence is initially encoded using Pose Guider and fused with multi-frame noise, followed by the Denoising UNet conducting the denoising process for video generation. The computational block of the Denoising UNet consists of Spatial-Attention, Cross-Attention, and Temporal-Attention, as illustrated in the dashed box on the right. The integration of reference image involves two aspects. Firstly, detailed features are extracted through ReferenceNet and utilized for Spatial-Attention. Secondly, semantic features are extracted through the CLIP image encoder for Cross-Attention. Temporal-Attention operates in the temporal dimension. Finally, the VAE decoder decodes the result into a video clip.}
\vspace{-0.3cm}
\label{fig:overview}
\end{figure*}

\section{Related Works}

\subsection{Diffusion Model for Image Generation }
In text-to-image research, diffusion-based methods\cite{dalle2,imagen,ldm,glide,ediffi,composer} have achieved significantly superior generation results, becoming the mainstream of research. To reduce computational complexity, Latent Diffusion Model\cite{ldm} proposes denoising in the latent space, striking a balance between effectiveness and efficiency. ControlNet\cite{controlnet} and T2I-Adapter\cite{t2iadapter} delve into the controllability of visual generation by incorporating additional encoding layers, facilitating controlled generation under various conditions such as pose, mask, edge and depth. Some studies further investigate image generation under given image conditions. IP-Adapter\cite{ip} enables diffusion models to generate image results that incorporate the content specified by a given image prompt. ObjectStitch\cite{objectstitch} and Paint-by-Example\cite{paint} leverage the CLIP\cite{clip} and propose diffusion-based image editing methods given image condition. TryonDiffusion\cite{tryondiffusion} applies diffusion models to the virtual apparel try-on task and introduces the Parallel-UNet structure. 

\subsection{Diffusion Model for Video Generation }
With the success of diffusion models in text-to-image applications, research in text-to-video has extensively drawn inspiration from text-to-image models in terms of model structure. 
Many studies\cite{text2videozero,fatezero,cogvideo,tuneavideo,rerender,gen1,followyourpose,makeavideo,vdm} explore the augmentation of inter-frame attention modeling on the foundation of text-to-image (T2I) models to achieve video generation.
Some works turn pretrained T2I models into video generators by inserting temporal layers. Video LDM\cite{align} proposes to first pretrain the model on images only and then train temporal layers on videos. AnimateDiff\cite{animatediff} presents a motion module trained on large video data which could be injected into most personalized T2I models without specific tuning. Our approach draws inspiration from such methods for temporal modeling. 

Some studies extend text-to-video capabilities to image-to-video.
VideoComposer\cite{videocomposer} incorporates images into the diffusion input during training as a conditional control. AnimateDiff\cite{animatediff} performs weighted mixing of image latent and random noise during denoising. VideoCrafter\cite{videocrafter1} incorporates textual and visual features from CLIP as the input for cross-attention. 
However, these approaches still face challenges in achieving stable human video generation, and the exploration of incorporating image condition input remains an area requiring further investigation.

\subsection{Diffusion Model for Human Image Animation }
Image animation\cite{fomm,mraa,ren2020deep,tpsmm,siarohin2019animating,zhang2022exploring,bidirectionally,everybody,liquid,editable}, aims to generate images or videos based on one or more input images. 
In recent research, the superior generation quality and stable controllability offered by diffusion models have led to their integration into human image animation. PIDM\cite{pidm} proposes texture diffusion blocks to inject the desired texture patterns into denoising for human pose transfer. LFDM\cite{LFDM} synthesizes an optical flow sequence in the latent space, warping the input image based on given conditions. LEO\cite{leo} represents motion as a sequence of flow maps and employs diffusion model to synthesize sequences of motion codes. DreamPose\cite{dreampose} utilizes pretrained Stable Diffusion model and proposes an adapter to model the CLIP and VAE image embeddings. DisCo\cite{disco} draws inspiration from ControlNet, decoupling the control of pose and background. Despite the incorporation of diffusion models to enhance generation quality, these methods still grapple with issues such as texture inconsistency and temporal instability in their results. Moreover, there is no method to investigate and demonstrate a more generalized capability in character animation.

\section{Methods}
We target at pose-guided image-to-video synthesis for character animation. Given a reference image describing the appearance of a character and a pose sequence, our model generates an animated video of the character. The pipeline of our method is illustrated in Fig.~\ref{fig:overview}. In this section, we first provide a concise introduction to Stable Diffusion in Sec~\ref{sec:sd}, which lays the foundational framework and network structure for our method. Then we provide a detailed explanation of the design specifics in Sec~\ref{sec:sd}. Finally, we present the training process in Sec~\ref{sec:train}.

\subsection{Preliminariy: Stable Diffusion}\label{sec:sd}
Our method is an extension of Stable Diffusion (SD), which is developed from Latent diffusion model (LDM). To reduce the computational complexity of the model, it introduces to model feature distributions in the latent space. SD develops an autoencoder\cite{vae,vqvae} to establish the implicit representation of images, which consists of an encoder $\mathcal E$ and a decoder $\mathcal D$. Given an image $\mathbf x$, the encoder first maps it to a latent representation: $\mathbf z$ = $\mathcal E$($\mathbf x$) and then the decoder reconstructs it: ${\mathbf x}_{recon}$ = $\mathcal D$($\mathbf z$). 

SD learns to denoise a normally-distributed noise $\epsilon$ to realistic latent $\mathbf z$. During training, the image latent $\mathbf z$ is diffused in $\mathnormal t$ timesteps to produce noise latent ${\mathbf z}_{t}$. And a denoising UNet is trained to predict the applied noise. The optimization process is defined as follow objective:

\begin{equation}
\label{eq1}
    {\mathbf L} = {\mathbb E}_{{\mathbf z}_{t},c,{\epsilon},t}({||{\epsilon}-{{\epsilon}_{\theta}}({\mathbf z}_{t},c,t)||}^{2}_{2})
\end{equation}

\noindent
where ${\epsilon}_{\theta}$ represents the function of the denoising UNet. $\mathnormal c$ represents the embeddings of conditional information. In original SD, CLIP ViT-L/14\cite{vit} text encoder is applied to represent the text prompt as token embeddings for text-to-image generation. The denoising UNet consists of four downsample layers , one middle layer and four upsample layers. A typical block within a layer includes three types of computations: 2D convolution, self-attention\cite{attention}, and cross-attention (terms as Res-Trans block). Cross-attention is conducted between text embedding and corresponding network feature. 

At inference, ${\mathbf z}_{T}$ is sampled from random Gaussian distribution with the initial timestep $\mathnormal T$ and is progressively denoised and restored to ${\mathbf z}_{0}$ via deterministic sampling process (e.g. DDPM\cite{denoising}, DDIM\cite{ddim}). 
In each iteration, the denoising UNet predicts the noise on the latent feature corresponding to each timestep $\mathnormal t$. 
Finally, ${\mathbf z}_{0}$ will be reconstructed by decoder $\mathcal D$ to obtain the generated image.

\subsection{Network Architecture}\label{sec:net}

\noindent
\textbf{Overview. }
Fig.~\ref{fig:overview} provides the overview of our method. The initial input to the network consists of multi-frame noise. The denoising UNet is configured based on the design of SD, employing the same framework and block units, and inherits the training weights from SD. Additionally, our method incorporates three crucial components: 1) ReferenceNet, encoding the appearance features of the character from the reference image; 2) Pose Guider, encoding motion control signals for achieving controllable character movements; 3) Temporal layer, encoding temporal relationship to ensure the continuity of character motion.

\noindent
\textbf{ReferenceNet. }
In text-to-video tasks, textual prompts articulate high-level semantics, necessitating only semantic relevance with the generated visual content. However, in image-to-video tasks, images encapsulate more low-level detailed features, demanding precise consistency in the generated results. In preceding studies focused on image-driven generation, most approaches\cite{ip,objectstitch,paint,dreampose,disco,videocrafter1} employ the CLIP image encoder as a substitute for the text encoder in cross-attention. However, this design falls short of addressing issues related to detail consistency. One reason for this limitation is that the input to the CLIP image encoder comprises low-resolution ($224{\times}224$) images, resulting in the loss of significant fine-grained detail information. Another factor is that CLIP is trained to match semantic features for text, emphasizing high-level feature matching, thereby leading to a deficit in detailed features within the feature encoding.

Hence, we devise a reference image feature extraction network named ReferenceNet. We adopt a framework identical to the denoising UNet for ReferenceNet, excluding the temporal layer. Similar to the denoising UNet, ReferenceNet inherits weights from the original SD, and weight update is conducted independently for each.
Then we explain the integration method of features from ReferenceNet into the denoising UNet. 
Specifically, as shown in Fig.~\ref{fig:overview}, we replace the self-attention layer with spatial-attention layer. Given a feature map ${x}_{1} {\in} {\mathbb R}^{{\mathnormal t}{\times}{\mathnormal h}{\times}{\mathnormal w}{\times}{\mathnormal c}}$ from denoising UNet and ${x}_{2} {\in} {\mathbb R}^{{\mathnormal h}{\times}{\mathnormal w}{\times}{\mathnormal c}}$ from ReferenceNet, we first copy ${x}_{2}$ by $\mathnormal t$ times and concatenate it with ${x}_{1}$ along $\mathnormal w$ dimension. Then we perform self-attention and extract the first half of the feature map as the output. This design offers two advantages: Firstly, ReferenceNet can leverage the pre-trained image feature modeling capabilities from the original SD, resulting in a well-initialized feature. Secondly, due to the essentially identical network structure and shared initialization weights between ReferenceNet and the denoising UNet, the denoising UNet can selectively learn features from ReferenceNet that are correlated in the same feature space.
Additionally, cross-attention is employed using the CLIP image encoder. Leveraging the shared feature space with the text encoder, it provides semantic features of the reference image, serving as a beneficial initialization to expedite the entire network training process.

A comparable design is ControlNet\cite{controlnet}, which introduces additional control features into the denoising UNet using zero convolution. However, control information, such as depth and edge, is spatially aligned with the target image, while the reference image and the target image are spatially related but not aligned. Consequently, ControlNet is not suitable for direct application. We will substantiate this in the subsequent experimental Section~\ref{ablation}.

While ReferenceNet introduces a comparable number of parameters to the denoising UNet, in diffusion-based video generation, all video frames undergo denoising multiple times, whereas ReferenceNet only needs to extract features once throughout the entire process. Consequently, during inference, it does not lead to a substantial increase in computational overhead.

\noindent
\textbf{Pose Guider. }
ControlNet\cite{controlnet} demonstrates highly robust conditional generation capabilities beyond text. Different from these methods, as the denoising UNet needs to be finetuned, we choose not to incorporate an additional control network to prevent a significant increase in computational complexity. Instead, we employ a lightweight Pose Guider. This Pose Guider utilizes four convolution layers ($4{\times}4$ kernels, $2{\times}2$ strides, using 16,32,64,128 channels, similar to the condition encoder in \cite{controlnet}) to align the pose image with the same resolution as the noise latent. Subsequently, the processed pose image is added to the noise latent before being input into the denoising UNet. The Pose Guider is initialized with Gaussian weights, and in the final projection layer, we employ zero convolution.

\noindent
\textbf{Temporal Layer. }
Numerous studies have suggested incorporating supplementary temporal layers into text-to-image (T2I) models to capture the temporal dependencies among video frames. This design facilitates the transfer of pretrained image generation capabilities from the base T2I model. Adhering to this principle, our temporal layer is integrated after the spatial-attention and cross-attention components within the Res-Trans block. The design of the temporal layer was inspired by AnimateDiff\cite{animatediff}. Specifically, for a feature map ${x} {\in} {\mathbb R}^{{\mathnormal b}{\times}{\mathnormal t}{\times}{\mathnormal h}{\times}{\mathnormal w}{\times}{\mathnormal c}}$, we first reshape it to ${x} {\in} {\mathbb R}^{({\mathnormal b}{\times}{\mathnormal h}{\times}{\mathnormal w}){\times}{\mathnormal t}{\times}{\mathnormal c}}$, and then perform temporal attention, which refers to self-attention along the dimension $\mathnormal t$. The feature from temporal layer is incorporated into the original feature through a residual connection. This design aligns with the two-stage training approach that we will describe in the following subsection.
The temporal layer is exclusively applied within the Res-Trans blocks of the denoising UNet. For ReferenceNet, it computes features for a single reference image and does not engage in temporal modeling. Due to the controllability of continuous character movement achieved by the Pose Guider, experiments demonstrate that the temporal layer ensures temporal smoothness and continuity of appearance details, obviating the need for intricate motion modeling.

\subsection{Training Strategy}\label{sec:train}
The training process is divided into two stages. In the first stage, training is performed using individual video frames. Within the denoising UNet, we temporarily exclude the temporal layer and the model takes single-frame noise as input. The ReferenceNet and Pose Guider are also trained during this stage. The reference image is randomly selected from the entire video clip. We initialize the model of the denoising UNet and ReferenceNet based on the pretrained weights from SD. The Pose Guider is initialized using Gaussian weights, except for the final projection layer, which utilizes zero convolution. The weights of the VAE's Encoder and Decoder, as well as the CLIP image encoder, are all kept fixed. The optimization objective in this stage is to enable the model to generate high-quality animated images under the condition of a given reference image and target pose.
In the second stage, we introduce the temporal layer into the previously trained model and initialize it using pretrained weights from AnimateDiff\cite{animatediff}. The input for the model consists of a 24-frames video clip. During this stage, we only train the temporal layer while fixing the weights of the rest of the network. 


\begin{figure*}[!t]
\begin{center}
	\setlength{\fboxrule}{0pt}
	\fbox{\includegraphics[width=1\linewidth]{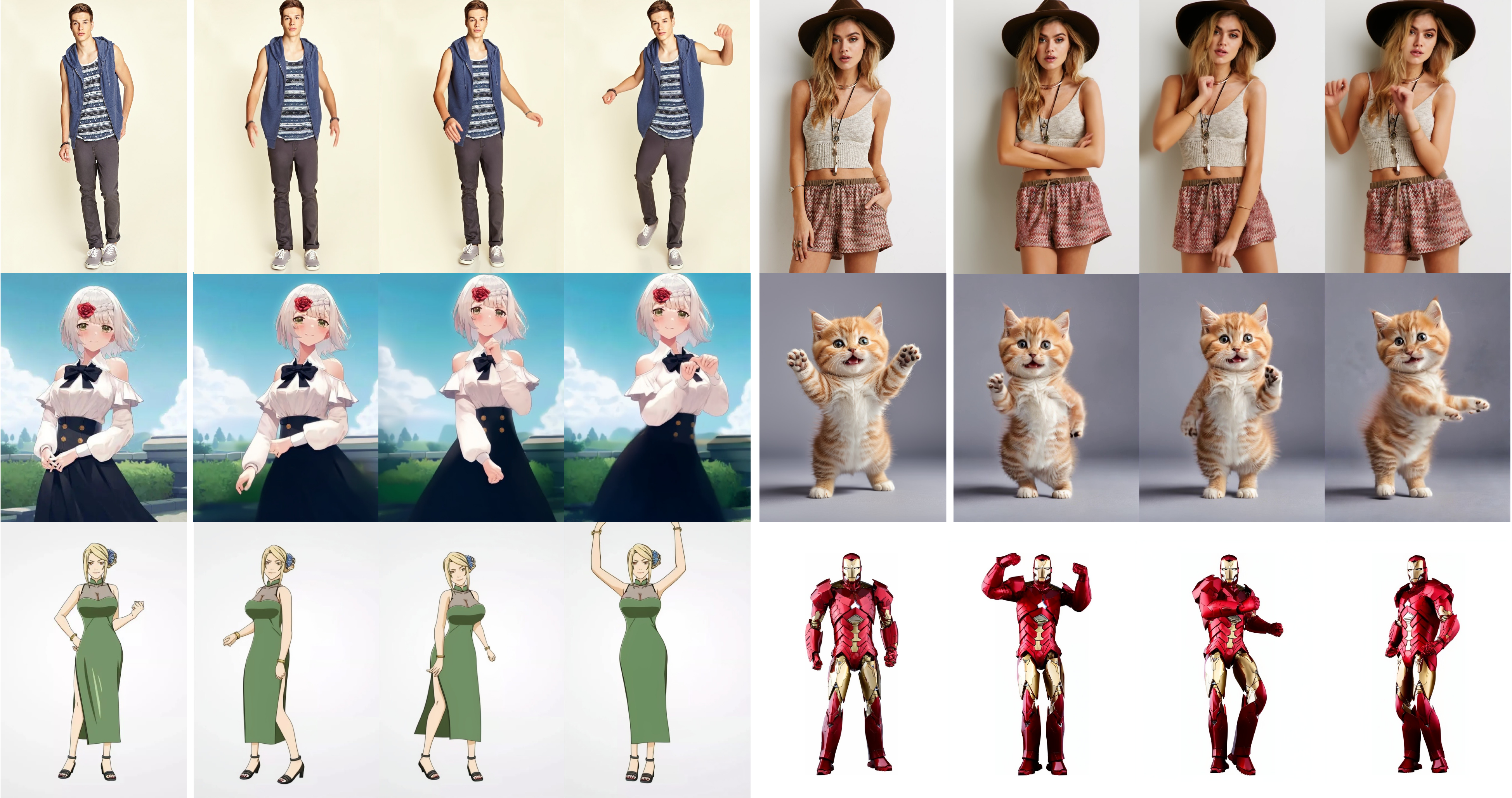}}
\end{center}
\vspace{-0.6cm}
\caption{Qualitative Results. Given a reference image (the leftmost image of each group), our approach demonstrates the ability to animate diverse characters, encompassing full-body human figures, half-length portraits, cartoon characters, and humanoid figures. The illustration showcases results with clear, consistent details, and continuous motion. }
\vspace{-0.2cm}
\label{fig:vis}
\end{figure*}

\section{Experiments}

\subsection{Implementations}

To demonstrate the applicability of our approach in animating various characters, we collect 5K character video clips from the internet to train our model.
We employ DWPose\cite{dwpose} to extract pose sequence of characters in the video, including body and hands, rendering it as pose skeleton images following OpenPose\cite{openpose}. Experiments are conducted on 4 NVIDIA A100 GPUs. In the first training stage, individual video frames are sampled, resized, and center-cropped to a resolution of $768{\times}768$. Training is conducted for 30,000 steps with a batch size of 64. In the second training stage, we train the temporal layer for 10,000 steps with 24-frame video sequences and a batch size of 4. Both learning rates are set to 1e-5. During inference, we rescale the length of the driving pose skeleton to approximate the length of the character's skeleton in the reference image and use a DDIM sampler for 20 denoising steps. We adopt the temporal aggregation method in \cite{edge}, connecting results from different batches to generate long videos. For fair comparison with other methods, we also train our model on three specific benchmarks (UBC fashion video dataset\cite{dwnet}, TikTok dataset\cite{tiktok} and Ted-talk dataset\cite{mraa}) without using additional data, as will be discussed in Section~\ref{com}. 

\subsection{Qualitative Results}
Fig.~\ref{fig:vis} demonstrates that our method can animate arbitrary characters, including full-body human figures, half-length portraits, cartoon characters, and humanoid characters. Our approach is capable of generating high-definition and realistic character details. It maintains temporal consistency with the reference images even under substantial motion and exhibits temporal continuity between frames. 

\subsection{Comparisons}\label{com}


To demonstrate the superiority of our approach, we evaluate its performance in three specific benchmarks: fashion video synthesis, human dance generation and talking gesture generation. We also conduct a baseline that combines Stable Diffusion, ControlNet, IP-Adapter\cite{ip} and AnimateDiff, named SD-I2V.
For quantitative assessment of image-level quality, SSIM\cite{ssim}, PSNR\cite{psnr} and LPIPS\cite{lpips} are employed. Video-level evaluation uses FVD\cite{fvd} metrics. 

\begin{figure}[!t]
\begin{center}
    \vspace{-0.3cm}
	\setlength{\fboxrule}{0pt}
	\fbox{\includegraphics[width=1\linewidth]{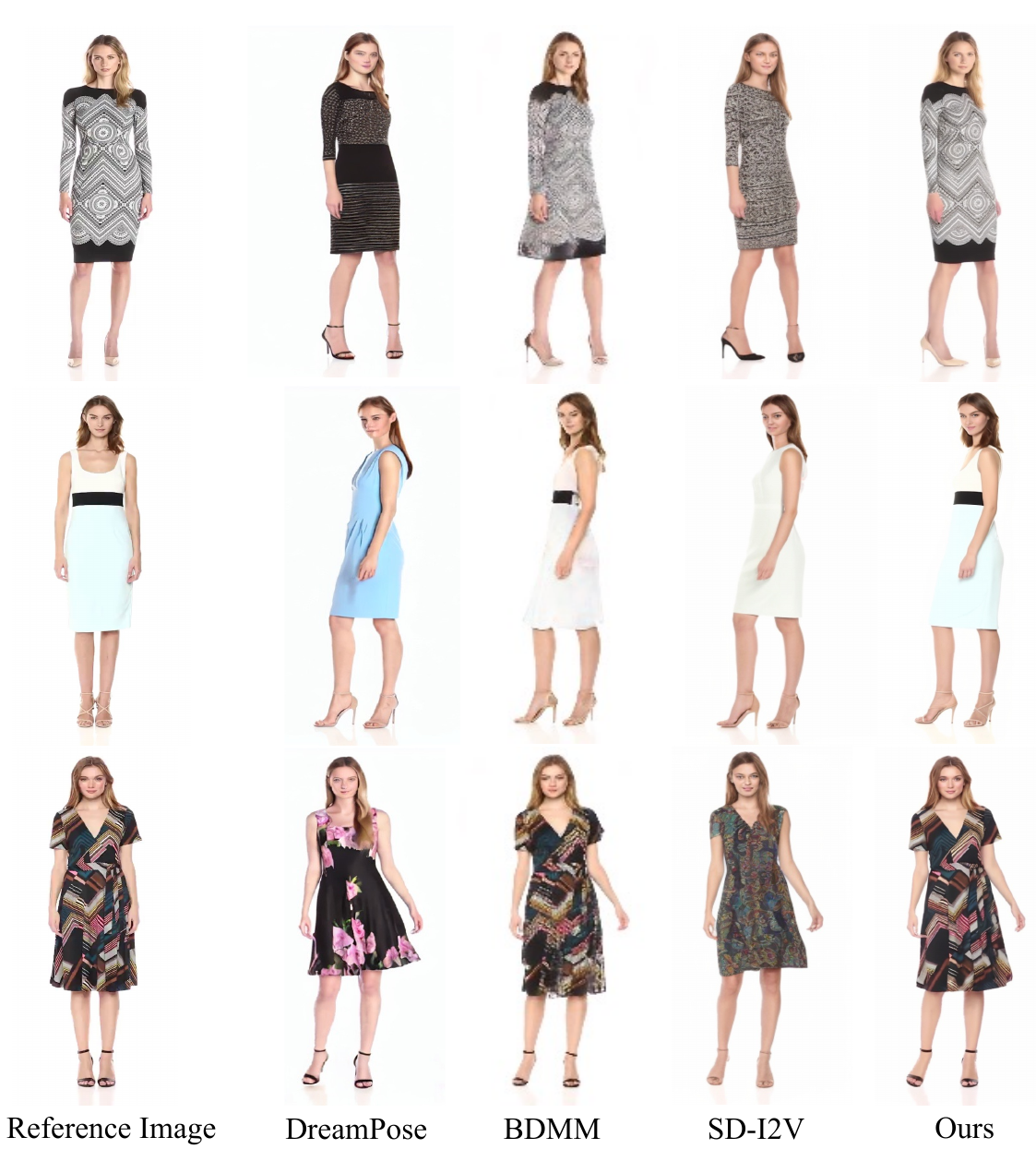}}
\end{center}
\vspace{-0.8cm}
\caption{Qualitative comparison for fashion video synthesis. Other methods exhibit shortcomings in preserving fine-textured details of clothing, whereas our method excels in maintaining exceptional detail features.}
\vspace{-0.2cm}
\label{fig:dreampose}
\end{figure}

\begin{table}
    \setlength{\tabcolsep}{5pt}
	\centering
	\begin{center}
\begin{tabular}{lccccc} 

\hline
    & SSIM $\uparrow$ & PSNR $\uparrow$ & LPIPS $\downarrow$  & FVD $\downarrow$ \\
 \hline
 MRAA\cite{mraa}     & 0.749 & - & 0.212    & 253.6 \\
 TPSMM\cite{tpsmm}    & 0.746 & - & 0.213  & 247.5 \\
 BDMM\cite{bidirectionally}     & 0.918 & 24.07 & 0.048  & 148.3 \\
DreamPose\cite{dreampose} & 0.885 & - & 0.068  & 238.7 \\
DreamPose* & 0.879 & 34.75 & 0.111  & 279.6 \\
SD-I2V & 0.894 & 36.01 & 0.095  & 175.4 \\
Ours & \textbf{0.931} & \textbf{38.49} & \textbf{0.044}  & \textbf{81.6} \\
 
\hline
\end{tabular}
\end{center}
    \vspace{-0.5cm}
	\caption{Quantitative comparison for fashion video synthesis. "Dreampose*" denotes the result without sample finetuning.}
    \vspace{-0.3cm}
	\label{table:ubc}
\end{table}

\noindent
\textbf{Fashion Video Synthesis. }
Experiments are conducted on the UBC fashion video dataset. 
The quantitative comparison is shown in Tab.~\ref{table:ubc}. Our result outperforms other methods, particularly exhibiting a significant lead in video metric. Qualitative comparison is shown in Fig.~\ref{fig:dreampose}. For fair comparison, we obtain results of DreamPose without sample finetuning using its open-source code. In the domain of fashion videos, there is a stringent requirement for fine-grained clothing details. However, other methods fail to maintain the consistency of clothing details and exhibit noticeable errors in terms of color and fine structural elements. In contrast, our method produces results that effectively preserve the consistency of clothing details.


\begin{figure}[!t]
\begin{center}
	\setlength{\fboxrule}{0pt}
	\fbox{\includegraphics[width=1\linewidth]{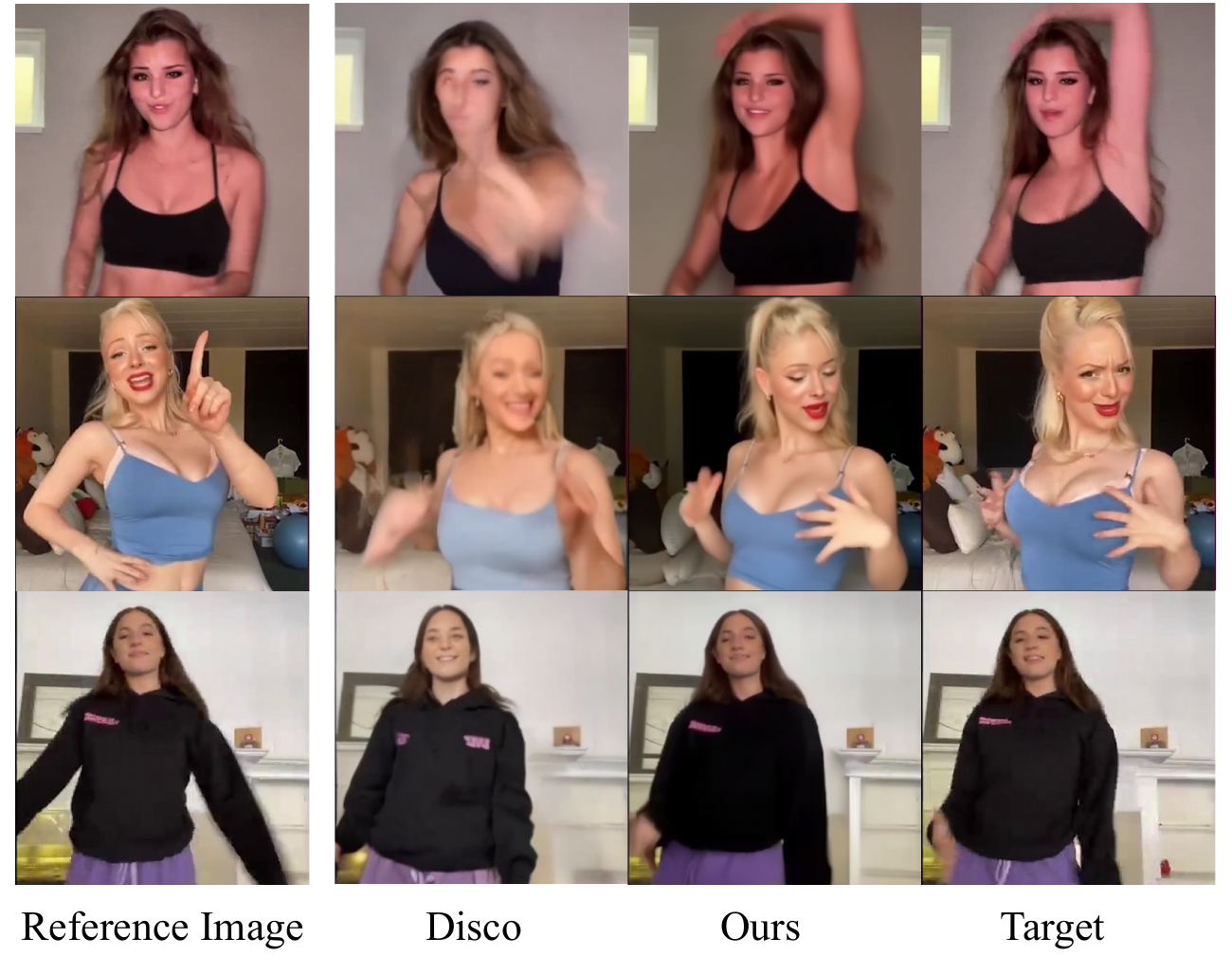}}
\end{center}
\vspace{-0.6cm}
\caption{Qualitative comparison between DisCo and our method. DisCo displays problems such as pose control errors, color inaccuracy, and inconsistent details. In contrast, our method demonstrates significant improvements in addressing these issues.}
\vspace{-0.2cm}
\label{fig:disco}
\end{figure}

\begin{table}
	\centering
	\begin{center}
\begin{tabular}{lccccc} 

\hline
    & SSIM $\uparrow$ & PSNR $\uparrow$ & LPIPS $\downarrow$ & FVD $\downarrow$ \\
 \hline
FOMM\cite{fomm}  & 0.648 & 29.01 & 0.335 & 405.2 \\
MRAA\cite{mraa} & 0.672 & 29.39 & 0.296 & 284.8 \\
TPSMM\cite{tpsmm} & 0.673 & 29.18 & 0.299 & 306.1 \\
Disco\cite{disco} & 0.668 & 29.03 & 0.292 & 292.8 \\
SD-I2V & 0.670 & 29.11 & 0.295 & 225.5 \\
Ours & \textbf{0.718} & \textbf{29.56} & \textbf{0.285} & \textbf{171.9} \\
 
\hline
\end{tabular}
\end{center}
    \vspace{-0.5cm}
	\caption{Quantitative comparison for human dance generation.}
    \vspace{-0.3cm}
	\label{table:tiktok}
\end{table}

\noindent
\textbf{Human Dance Generation. }
We conduct experiments on the TikTok dataset. 
We conduct a quantitative comparison presented in Tab.~\ref{table:tiktok}, and our method achieves the best results.
For enhanced generalization, DisCo incorporates human attribute pre-training, utilizing a large number of image pairs for model pre-training. In contrast, our training is exclusively conducted on the TikTok dataset, yielding results superior to DisCo. 
We present qualitative comparison with DisCo in Fig.~\ref{fig:disco}. 
During intricate dance sequences, our model stands out in maintaining visual continuity throughout the motion and exhibits enhanced robustness in handling diverse character appearances.

\begin{figure}[!t]
\begin{center}
	\setlength{\fboxrule}{0pt}
	\fbox{\includegraphics[width=1\linewidth]{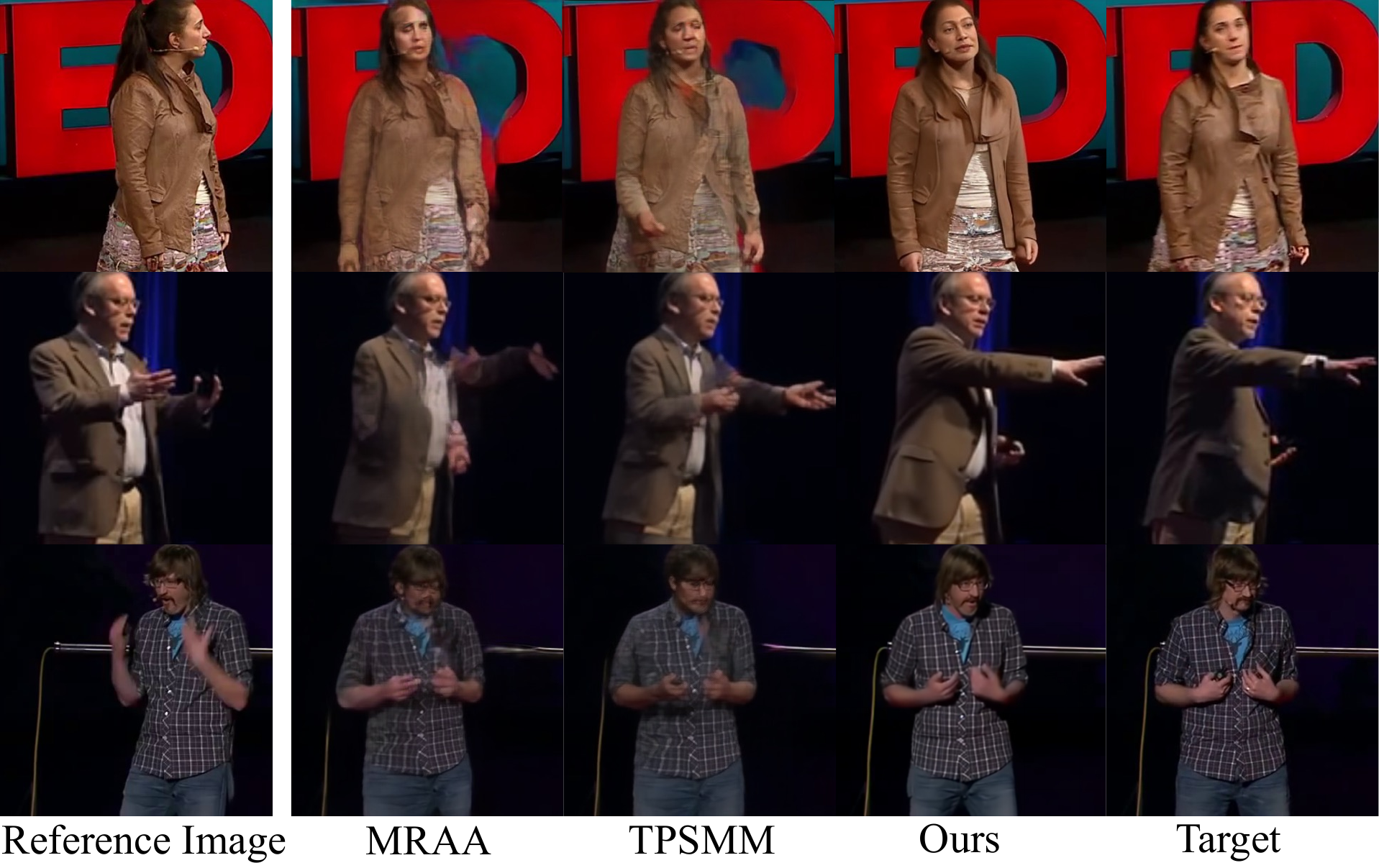}}
\end{center}
\vspace{-0.6cm}
\caption{Qualitative comparison on Ted-talk dataset. Our model is capable of generating more accurate and clear results.}
\vspace{-0.2cm}
\label{fig:rebuttal_fig}
\end{figure}

\begin{table}
	\centering
	\begin{center}
\begin{tabular}{lccccc} 

\hline
    & SSIM $\uparrow$ & PSNR $\uparrow$ & LPIPS $\downarrow$ & FVD $\downarrow$ \\
 \hline
MRAA\cite{mraa} & 0.826 & 33.86 & 0.160 & 82.8 \\
TPSMM\cite{tpsmm} & 0.830 & 33.81 & \textbf{0.157} & 80.7 \\
Disco\cite{disco} & 0.754 & 31.25 & 0.193 & 223.5 \\
SD-I2V  & 0.773 & 32.11 & 0.179 & 158.3 \\
Ours & \textbf{0.832} & \textbf{33.91} & 0.159 & \textbf{80.5} \\
 
\hline
\end{tabular}
\end{center}
    \vspace{-0.5cm}
	\caption{Quantitative comparison on Ted-talk dataset.}
    \vspace{-0.3cm}
	\label{table:tedtalk}
\end{table}

\noindent
\textbf{Talking Gesture Generation. }
We also evaluate our method on Ted-talk dataset. 
Results are shown in Fig \ref{fig:rebuttal_fig} and Tab \ref{table:tedtalk}. Our approach significantly outperforms DisCo and SD-I2V. MRAA and TPSMM employ GT images as driving signals (video reconstruction), while we achieve better results using only pose information. On other two evaluated benchmarks (UBC with more intricate clothing textures and TikTok with more complex human movements), the performance of MRAA and TPSMM falls far behind our method.

\begin{figure}[!t]
\begin{center}
	\setlength{\fboxrule}{0pt}
	\fbox{\includegraphics[width=1\linewidth]{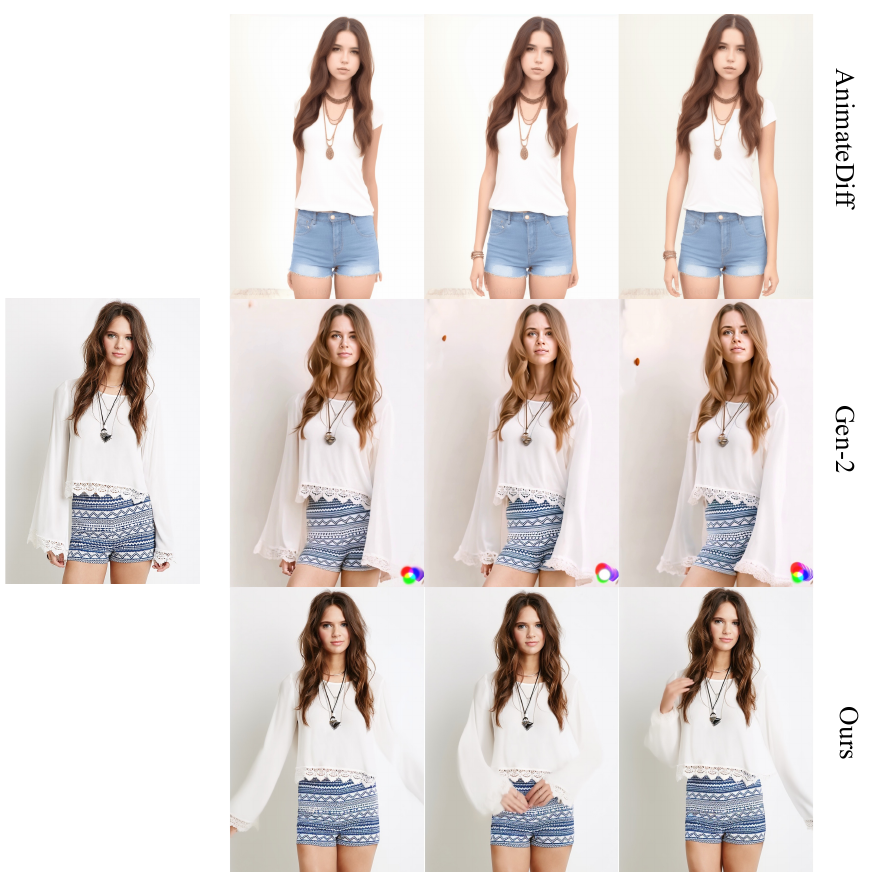}}
\end{center}
\vspace{-0.6cm}
\caption{Qualitative comparison with image-to-video methods, which struggle to generate substantial character movements and face challenges in maintaining long-term appearance consistency.}
\vspace{-0.2cm}
\label{fig:i2v}
\end{figure}

\noindent
\textbf{General Image-to-Video Methods. }
Currently, numerous studies propose video diffusion models with strong generative capabilities based on large-scale training data. We select two of the most well-known and effective image-to-video methods for comparison: AnimateDiff\cite{animatediff} and Gen-2\cite{gen1}. As these two methods do not perform pose control, we only compare their ability to maintain the appearance fidelity to the reference image. As depicted in Fig.~\ref{fig:i2v}, current image-to-video methods face challenges in generating substantial character movements and struggle to maintain long-term appearance consistency in videos, thus hindering effective support for consistent character animation.

\subsection{Ablation study}\label{ablation}

\begin{figure}[!t]
\begin{center}
    \vspace{-0.3cm}
	\setlength{\fboxrule}{0pt}
	\fbox{\includegraphics[width=1\linewidth]{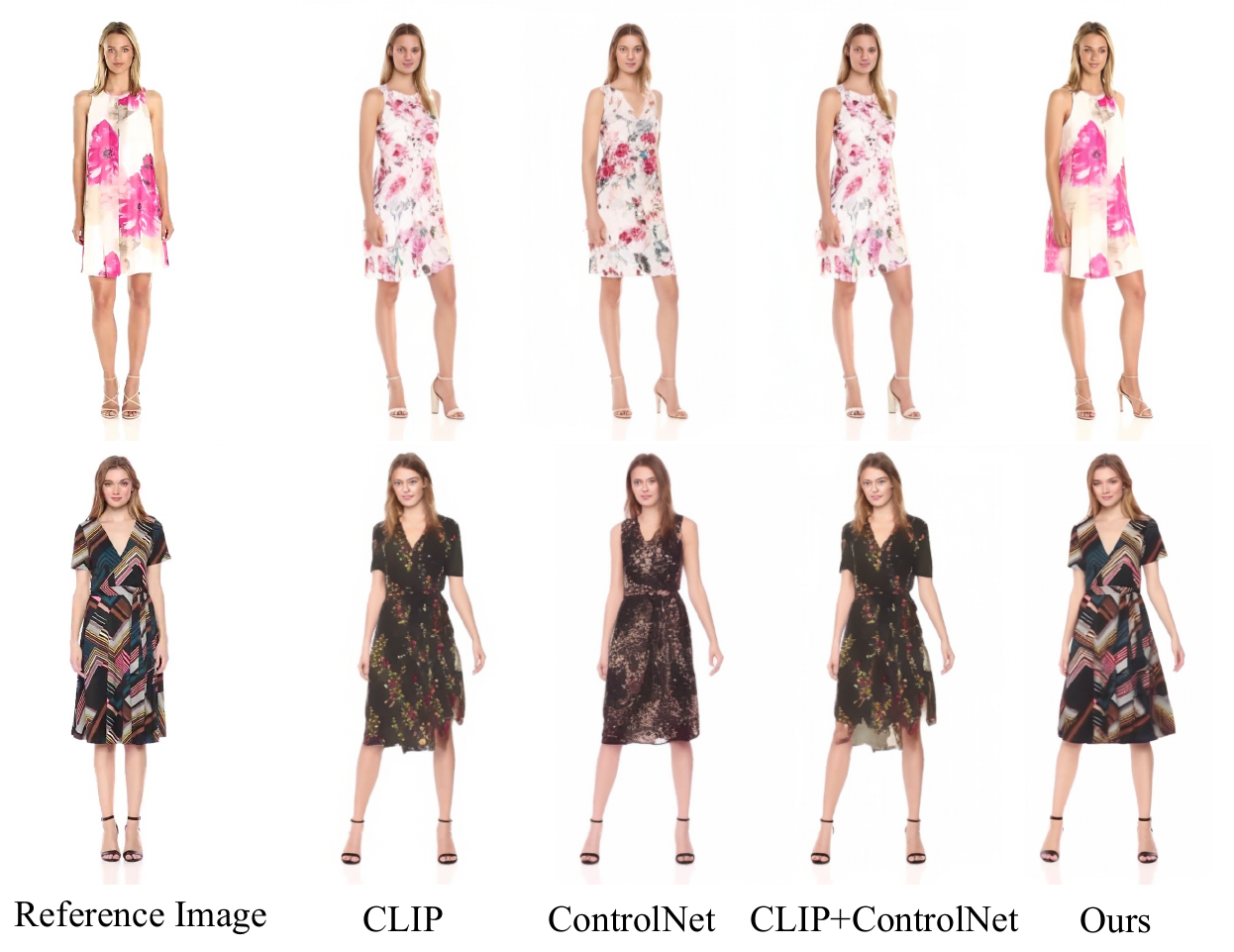}}
\end{center}
\vspace{-0.6cm}
\caption{Ablation study of different design. ReferenceNet ensures consistent preservation of details in character's appearance.}
\vspace{-0.2cm}
\label{fig:ablation}
\end{figure}

\noindent
\textbf{Image Condition Modeling. }
To demonstrate the effectiveness of our image condition modeling, we explore alternative designs, including 1) using only the CLIP image encoder to represent reference image features without integrating ReferenceNet, 2) initially finetuning SD and subsequently training ControlNet with the reference image. 3) integrating the above two designs. 
Experiments are conducted on UBC fashion video dataset. 
As shown in Fig.~\ref{fig:ablation}, visualizations illustrate that ReferenceNet outperforms the other three designs. Solely relying on CLIP features as reference image features can preserve image similarity but fails to fully transfer details. ControlNet does not enhance results as its features lack spatial correspondence, rendering it inapplicable. Quantitative results are also presented in Tab.~\ref{table:ablation}, demonstrating the superiority of our design.

\noindent
\textbf{Details of ReferenceNet Design. }
To demonstrate the effectiveness of our ReferenceNet design, we conduct experiments: 1) Replacing UNet (SD weights) with ResNet (ImageNet weights). 2) Replacing spatial-attention with feature-concatenation. Quantitative result is shown in Tab \ref{table:ablation_rebuttal}. Our design achieves the optimal performance. Experiment 1) shows the necessity of utilizing SD weights. There exists a certain gap between ImageNet's image features and the implicit features in SD. Utilizing features from SD enhances the integration of conditioning information within the same feature space during the generation process. Experiment 2) demonstrates the necessity of spatial-attention which enables the denoising UNet to effectively integrate detailed image features from ReferenceNet.

\begin{table}
    \setlength{\tabcolsep}{4pt}
	\centering
	\begin{center}
\begin{tabular}{lcccc} 

\hline
    & SSIM $\uparrow$ & PSNR $\uparrow$ & LPIPS $\downarrow$  & FVD $\downarrow$ \\
 \hline
 CLIP     & 0.897 & 36.09 & 0.089    & 208.5 \\
 ControlNet    & 0.892 & 35.89 & 0.105  & 213.9 \\
CLIP+ControlNet & 0.898 & 36.03 & 0.086  & 205.4 \\
Ours & \textbf{0.931} & \textbf{38.49} & \textbf{0.044}  & \textbf{81.6} \\
 
\hline
\end{tabular}
\end{center}
    \vspace{-0.5cm}
	\caption{Ablation study for image condition modeling.}
    \vspace{-0.3cm}
	\label{table:ablation}
\end{table}

\begin{table}
    \setlength{\tabcolsep}{4pt}
	\centering
	\begin{center}
\begin{tabular}{lcccc} 

\hline
    & SSIM $\uparrow$ & PSNR $\uparrow$ & LPIPS $\downarrow$  & FVD $\downarrow$ \\
 \hline
 ImageNet weights     & 0.901 & 36.21 & 0.084    & 165.4 \\
 Feature-concat    & 0.909 & 36.53 & 0.071  & 132.8 \\
Ours & \textbf{0.931} & \textbf{38.49} & \textbf{0.044}  & \textbf{81.6} \\
 
\hline
\end{tabular}
\end{center}
    \vspace{-0.5cm}
	\caption{Ablation study for ReferenceNet design.}
    \vspace{-0.3cm}
	\label{table:ablation_rebuttal}
\end{table}

\noindent
\textbf{Temporal Modeling. }
We conduct two experiments to assess the effectiveness of proposed temporal modeling method: 
1) do not apply temporal layer, directly concatenating images temporally to create a video. 
2) do not apply the two-stage training, directly training the entire network. 
Quantitative results are presented in Tab.~\ref{table:temporal}. The absence of temporal layer results in noticeable texture sticking and inter-frame jitter, leading to a significant decrease in FVD metric. When the two-stage training is not employed, metrics related to image quality experience a decline. We attribute this to the fact that, when optimizing over multiple frames simultaneously, the network tends to focus more on the overall temporal visual coherence, thereby weakening attention to the details of each individual frame. The adoption of the two-stage training method ensures both the quality of generated video frames and temporal smoothness.

\begin{table}
    \setlength{\tabcolsep}{2pt}
	\centering
	\begin{center}
\begin{tabular}{lcccc} 

\hline
    & SSIM $\uparrow$ & PSNR $\uparrow$ & LPIPS $\downarrow$  & FVD $\downarrow$ \\
 \hline
 w/o Temporal Layer     & 0.925 & 38.28 & 0.049    & 176.7 \\
 w/o Two Stage Training    & 0.917 & 38.01 & 0.056  & 89.3 \\
Ours & \textbf{0.931} & \textbf{38.49} & \textbf{0.044}  & \textbf{81.6} \\
 
\hline
\end{tabular}
\end{center}
    \vspace{-0.5cm}
	\caption{Ablation study of temporal modeling.}
    \vspace{-0.3cm}
	\label{table:temporal}
\end{table}





\section{Discussion and Conclusion}

\noindent
\textbf{Limitations. }
Our model may struggle to generate stable results for hand movements, sometimes leading to distortions and motion blur. 
Besides, since images provide information from only one perspective, generating unseen parts during character movement is an ill-posed problem which encounters potential instability. 
Third, due to the utilization of DDPM, our model exhibits a lower operational efficiency compared to non-diffusion-model-based methods.

\noindent
\textbf{Potential Impact. }
The proposed method may be used to produce fake videos of individuals, which can be detected using some face anti-spoofing techniques\cite{anti_color,anti_deep,anti_search}. 

\noindent
\textbf{Conclusion. }
In this paper, we present \textit{Animate Anyone}, a framework capable of transforming character photographs into animated videos controlled by a desired pose sequence.
We propose ReferenceNet, which genuinely preserves intricate character appearances and we also achieve efficient pose controllability and temporal continuity. Our approach not only applies to general character animation but also outperforms existing methods. 


{
    \small
    \bibliographystyle{ieeenat_fullname}
    \bibliography{main}
}


\end{document}